\documentclass{article}

\usepackage[preprint]{neurips_2024}

\usepackage[utf8]{inputenc} % allow utf-8 input
\usepackage[T1]{fontenc}    % use 8-bit T1 fonts
\usepackage{hyperref}       % hyperlinks
\usepackage{url}            % simple URL typesetting
\usepackage{booktabs}       % professional-quality tables
\usepackage{amsfonts}       % blackboard math symbols
\usepackage{nicefrac}       % compact symbols for 1/2, etc.
\usepackage{microtype}      % microtypography
\usepackage[table]{xcolor}
\definecolor{NewColor1}{RGB}{255, 102, 0}
\definecolor{NewColor2}{RGB}{0, 150, 255}
\definecolor{cellpurple}{RGB}{210, 210, 240} % Light purple for S1
\definecolor{cellblue}{RGB}{200, 230, 245}   % Light blue for S2/S3
\definecolor{cellgreen}{RGB}{190, 240, 200}  % Light green for Totals

\usepackage{graphicx}
\usepackage{amsmath}
\usepackage{booktabs}
\usepackage{cite}
\usepackage{subcaption}

\title{RoCo Challenge at AAAI 2026: Benchmarking \textcolor{NewColor2}{Ro}botic \textcolor{NewColor1}{Co}llaborative Manipulation for Assembly Towards Industrial Automation}

\author{%
  \textbf{Haichao Liu$^1$, Yuheng Zhou$^1$, Zhenyu Wu$^2$, Ziheng Ji$^1$, Ziyu Shan$^1$, Qianzhun Wang$^1$,} \\
  \textbf{Ruixuan Liu$^3$, Zhiyuan Yang$^4$, Yejun Gu$^4$, Shalman Khan$^4$, Shijun Yan$^4$, Jun Liu$^4$,} \\
  \textbf{Haiyue Zhu$^4$, Changliu Liu$^3$, Jianfei Yang$^1$, Jingbing Zhang$^4$, and Ziwei Wang$^{1}$}\thanks{Corresponding author.} \vspace{0.3cm}\\
  $^1$Nanyang Technological University, Singapore \\
  $^2$Beijing University of Posts and Telecommunications, Beijing, China \\
  $^3$Carnegie Mellon University, Pittsburgh, USA \\
  $^4$Agency for Science, Technology and Research (A*STAR), Singapore \\
  \texttt{} \\
}

\begin{document}

\maketitle

\begin{abstract}
Embodied Artificial Intelligence (EAI) is rapidly developing, gradually subverting previous autonomous systems' paradigms from isolated perception to integrated, continuous action. This transition is highly significant for industrial robotic manipulation, promising to free human workers from repetitive, dangerous daily labor. To benchmark and advance this capability, we introduce the Robotic Collaborative Assembly Assistance (RoCo) Challenge with a dataset towards simulation and real-world assembly manipulation. Set against the backdrop of human-centered manufacturing, this challenge focuses on a high-precision planetary gearbox assembly task, a demanding yet highly representative operation in modern industry.
Built upon a self-developed data collection, training, and evaluation system in Isaac Sim, and utilizing the Galaxea R1 Lite dual-arm robot for real-world deployment, the challenge operates in two phases. The Simulation Round defines fine-grained task phases for step-wise scoring to handle the long-horizon nature of the assembly. The Real-World Round mirrors this evaluation with physical gearbox components and high-quality teleoperated datasets. The core tasks require assembling an epicyclic gearbox from scratch, including mounting three planet gears, a sun gear, and a ring gear. 
Attracting over 60 teams and 170+ participants from more than 10 countries, the challenge yielded highly effective solutions, most notably ARC-VLA and RoboCola. Results demonstrate that a Specialist + Generalist dual-model framework for long-horizon multi-task learning is highly effective, and the strategic utilization of recovery-from-failure curriculum data is a critical insight for successful deployment. This report outlines the competition setup, task design, evaluation approach, key findings, and future directions for industrial EAI. Our dataset, CAD files of industrial objects, code, and evaluation results can be found at: \textcolor{blue}{https://rocochallenge.github.io/RoCo2026/}.
\end{abstract}

\section{Introduction}

Embodied Artificial Intelligence (EAI) is currently undergoing a transformative shift, transitioning from isolated digital twins to integrated agents capable of executing complex physical tasks in unstructured environments~\citep{eaisurvey}. This evolution is gradually subverting traditional autonomous systems' paradigms, moving from rigid, pre-programmed perception-action sequences toward adaptive, foundation-model-driven architectures~\citep{robofoundationsurvey}. Within the broader scope of EAI, robotic manipulation for industrial use cases remains a critical frontier~\citep{manusurvey}. Automating intricate assembly tasks not only promises to enhance manufacturing efficiency but also serves a vital socio-economic role by freeing human workers from repetitive, ergonomically taxing, and potentially hazardous daily labors~\citep{manureview}.

A number of recent benchmarks have been proposed to study embodied manipulation, spanning both simulation and real-world settings. In simulation, benchmarks such as \textbf{LIBERO}~\citep{libero} provide procedurally generated lifelong manipulation tasks designed to study knowledge transfer and continual learning across task suites. \textbf{ManiSkill~3}~\citep{maniskill3} introduces a GPU-parallelized robotics simulator that enables large-scale training on diverse contact-rich manipulation environments with high simulation throughput. \textbf{CALVIN}~\citep{calvin} focuses on language-conditioned long-horizon manipulation, where robots must compose multiple skills to execute sequential tasks based on natural language instructions. \textbf{RoboTwin~2.0}~\citep{robotwin2} further emphasizes scalable synthetic data generation and evaluation for robust bimanual manipulation through extensive domain randomization. 

In real-world settings, several benchmarks have also been developed to study physical robot learning. \textbf{LeRobot}~\citep{lerobot} provides an open-source framework and infrastructure for collecting and training on real-world robotic datasets, aiming to standardize the robot learning stack across hardware platforms. \textbf{FurnitureBench}~\citep{furniturebench} focuses on long-horizon furniture assembly tasks and provides standardized hardware setups together with large teleoperated demonstration datasets to facilitate reproducible evaluation. The \textbf{Functional Manipulation Benchmark (FMB)}~\citep{fmb} studies generalizable robotic learning through procedurally generated 3D-printed objects and functional manipulation tasks such as grasping, repositioning, and assembly.

Despite these advances, existing benchmarks primarily focus on household manipulation, tabletop tasks, or synthetic environments, and rarely capture the precision requirements, structured workflows, and collaborative manipulation characteristics of real industrial assembly processes. In particular, high-precision mechanical assemblies such as gearbox construction require coordinated bimanual manipulation, strict tolerance control, and robust recovery from intermediate failures, aspects that remain underexplored in current benchmarks. The RoCo Challenge is designed to address this gap by introducing a high-fidelity benchmark centered on industrial gearbox assembly, bridging simulation and real-world evaluation under a unified framework.

Technically, achieving human-level dexterity and reliability in industrial settings presents formidable technical challenges, which are further compounded by the absence of representative benchmarks for industrial assembly manipulation. First, \textbf{long-horizon manipulation} requires the robot to maintain a coherent global plan while executing a series of interdependent sub-tasks, where a single failure in an early stage can propagate throughout the entire workflow~\citep{tampsurvey}. Second, \textbf{accurate interaction with objects}, specifically in contact-rich tasks like gearbox assembly, demands sub-millimeter precision that is often difficult to achieve with visual feedback alone~\citep{contactrichsurvey}. Third, the transition to \textbf{coordinated bimanual manipulation} introduces increased degrees of freedom and the need for sophisticated spatial reasoning~\citep{bimanual}. Finally, a truly "human-centric" system must possess \textbf{recovery from failure} capabilities, ensuring that the robot can detect anomalies, such as a misplaced part or an incorrect component, and autonomously correct its state to ensure robust industrial performance~\citep{recoveryfailures}.

Current technical approaches to these challenges generally fall into two categories. \textbf{Modularized perception-action paradigms} decompose the problem into discrete stages: perception and 3D reconstruction, task and motion planning (TAMP), and foundation pose estimation for grasp perception~\citep{robodexvlm}. While interpretable, these pipelines are often brittle to sensor noise and calibration errors. Conversely, \textbf{end-to-end data-driven approaches} have gained traction, ranging from high-frequency visuomotor policies like Diffusion Policy (DP)~\citep{dp} and Action Chunking Transformer (ACT)~\citep{act} to large-scale Vision-Language-Action (VLA) models such as RT-2~\citep{rt2}, $\pi_0$~\citep{pi0}, and GR-00T~\citep{gr00t}. 

However, deploying these models in real-world factories uncovers significant hurdles~\citep{manusurvey, eaisurvey}:
\begin{itemize}
    \item \textbf{Out-of-Distribution (OOD) Generalization:} Models often struggle when encountering environment configurations or lighting conditions not present in the training data.
    \item \textbf{Data Efficiency:} Learning complex skills for high-precision tasks typically requires vast amounts of data, which is expensive to collect in physical industrial settings.
    \item \textbf{Robustness to Noise:} Models must learn to extract actionable features from noisy, multimodal real-world data without the luxury of high-fidelity tactile or force-sensing modalities.
    \item \textbf{Real-time Feasibility:} The computational demand of large-scale models must be balanced against the constraints of edge hardware to ensure low-latency deployment.
\end{itemize}

To systematically address these challenges, we introduced the \textbf{RoCo Challenge} as part of the International Workshop on Addressing Challenges and Opportunities in Human-Centric Manufacturing (HCM) at \textbf{AAAI 2026}. The challenge serves as a bridge between the global research community and industrial application, providing a high-fidelity Simulation Track based on NVIDIA Isaac Sim~\citep{isaacsim} for rapid iteration and an Onsite Track utilizing the \textbf{Galaxea R1 Lite} dual-arm platform~\citep{r1lite} for real-world validation. 

By benchmarking state-of-the-art solutions on an epicyclic gearbox assembly task, we aim to push the boundaries of EAI in industrial conditions, setting new standards for collaborative manipulation. This report outlines the competition setup, task design, and evaluation approach, and provides an in-depth analysis of the key findings and future directions identified by the participating teams to catalyze the next generation of smart manufacturing solutions.

\section{Competition Structure, Rules and Results}
The RoCo Challenge was meticulously designed to bridge the gap between high-fidelity simulation and physical industrial deployment. The competition was divided into two distinct but interconnected tracks: the Simulation Track, which emphasized algorithmic iteration and data efficiency, and the Onsite Track, which focused on the robustness of real-world deployment on standardized hardware.
\subsection{Simulation Track: Human-in-the-Loop by State Initialization}
The simulation phase utilized a self-developed environment built upon the NVIDIA Isaac Sim platform. This environment provided teams with a digital twin of the assembly bench and the Galaxea R1 robot, allowing for large-scale data collection and policy training.
\subsubsection{Task Definitions and Data Collection}
\begin{figure}[t]
    \centering
    \scalebox{1}[-1]{\includegraphics[width=0.5\columnwidth]{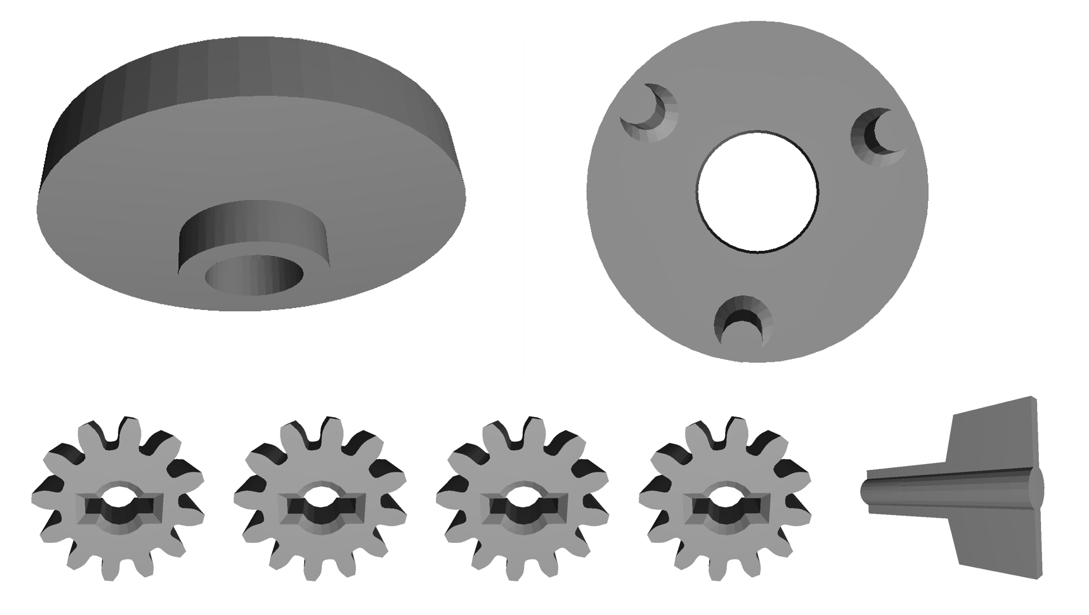}}
    \caption{The simplified planetary gear reducer used in the RoCo Challenge, consisting of a sun gear, three planet gears, one finned key shaft, a ring gear component, and a planet carrier with three pins. The assembly task requires precise placement of these components within a planet carrier.}
    \label{fig:gear_reducer_flipped}
\end{figure}
To evaluate the long-horizon nature of gearbox assembly, leveraging the simplified planary gear reducer shown in Figure~\ref{fig:gear_reducer_flipped} the simulation track defined three representative scenarios:
\begin{itemize}
    \item \textbf{Task 1 (Assembly from Scratch):} The robot begins with an empty planet carrier and must sequentially install three planet gears, the sun gear, and the ring gear.
    \item \textbf{Task 2 (Resume from Partial State):} The robot is initialized in a state where some components have already been assembled (simulating a handover from a "virtual human operator"). The agent must perceive the current state and continue the sequence correctly.
    \item \textbf{Task 3 (Error Detection and Recovery):} Semantic errors, such as the placement of an incorrect gear size, are injected into the initial state. The robot must identify the discrepancy, remove the faulty part, and resume the correct assembly flow.
\end{itemize}

\begin{figure}[t]
     \centering
     \begin{subfigure}[b]{\linewidth}
         \centering
         \includegraphics[width=\linewidth]{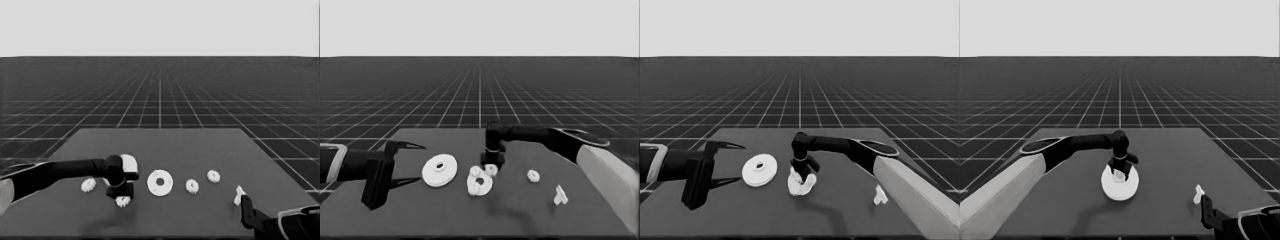}
         \caption{Task 1: Scratch assembly}
         \label{fig:sim1}
     \end{subfigure}
     \hfill
     \begin{subfigure}[b]{\linewidth}
         \centering
         \includegraphics[width=\linewidth]{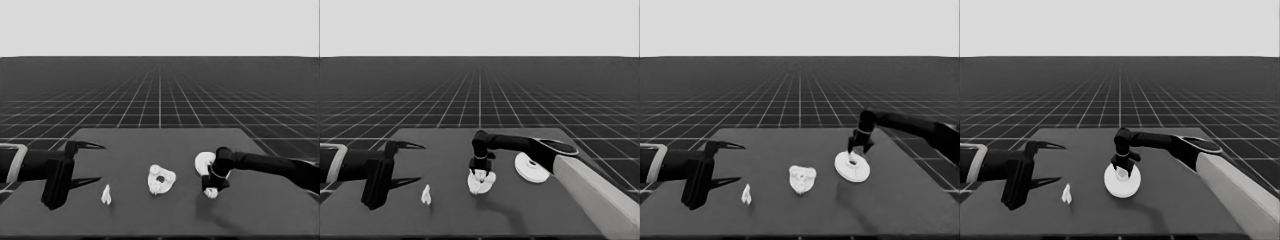}
         \caption{Task 2: Partial state}
         \label{fig:sim2}
     \end{subfigure}
     \hfill
     \begin{subfigure}[b]{\linewidth}
         \centering
         \includegraphics[width=\linewidth]{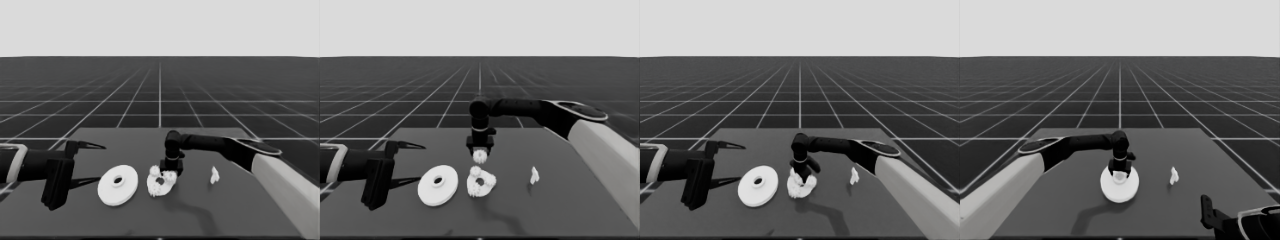}
         \caption{Task 3: Gear recovery}
         \label{fig:sim3}
     \end{subfigure}
        \caption{Simulation environment demonstrations for the three RoCo Challenge tasks. The stages mirror the real-world setup: (a) initial assembly, (b) completion from partial states, and (c) dynamic error recovery.}
        \label{fig:sim_tasks_all}
\end{figure}

As shown in Figure~\ref{fig:sim_tasks_all}, In task 1 (Figure~\ref{fig:sim1}) the robot assembles from scratch: pick up the planet gears and insert them onto the pins on the planet carrier; pick up the sun gear and insert it in the center of planet gears; finally pick up the ring gear, and place it onto the gears. In task 2 (Figure~\ref{fig:sim2}) the robot starts from a partial state where three planetary gears are already in place, and it only needs to pick up and mount the sun gear and the ring gear. In task 3 (Figure~\ref{fig:sim3}) the error is defined that the sun gear is placed on top of one planet gear by mistake, and the robot needs to pick the sun gear from that wrong place to proceed with normal assembly.

\begin{figure}[t]
    \centering
    \includegraphics[width=0.8\columnwidth]{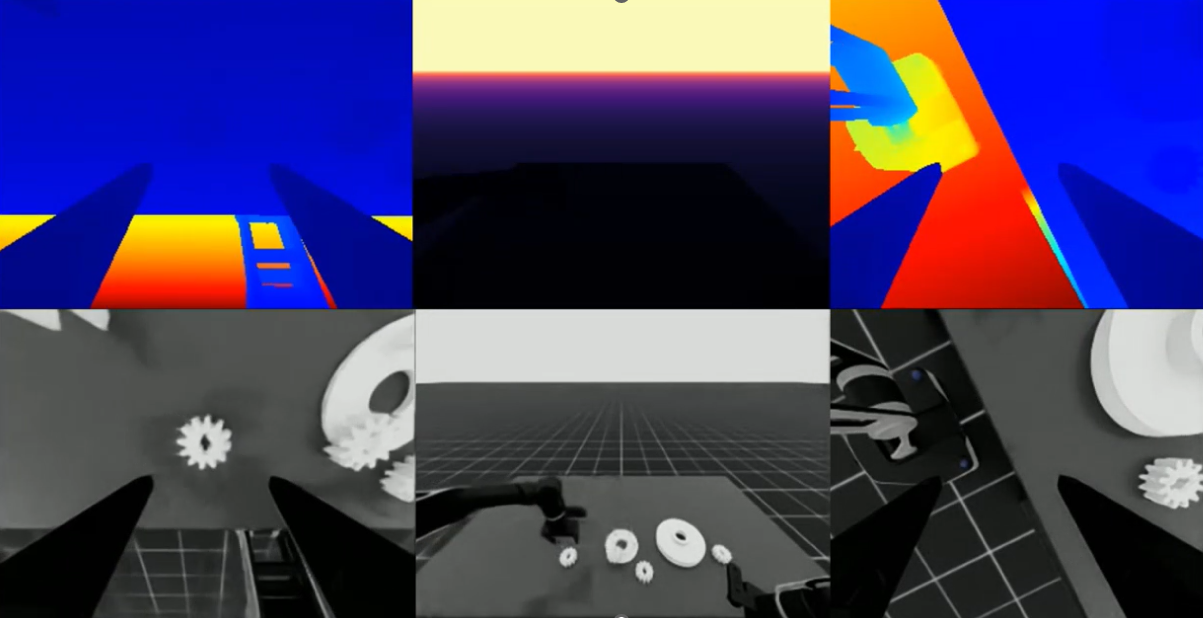}
    \caption{Visualization of the data frame for the simulation track. The dataset includes synchronized multi-modal observations (RGB, Depth, and proprioception) and action streams, enabling comprehensive training for both visuomotor imitation learning and VLA models.}
    \label{fig:sim_data_frame}
\end{figure}

\subsubsection{Evaluation and Scoring}
A single model was required to handle all three scenarios. Success was measured using a normalized score ($S \in [0, 1]$) based on the number of correctly assembled parts and the accuracy of error recovery. The final simulation score was calculated using a weighted scheme:$$S_{sim} = \frac{4 \cdot S_1 + 2 \cdot S_2 + 4 \cdot S_3}{10}$$To ensure fairness and prevent environment-specific overfitting, the final evaluation was executed on a private workstation with randomized (but consistent across teams) lighting, textures, and gearbox colors.

\subsubsection{Baseline and Data Provision}
To facilitate entry and encourage innovation, we provided baseline implementations of a VLA model ($\pi_{0.5}$~\citep{pi05}) and an ACT model~\citep{act}, trained on a subset of the provided dataset. These baselines served as reference points for participants to improve upon. The detailed deployment instructions of the baselines, including code and hyperparameters, were made available on the official competition website.

In terms of data, teams were given access to a comprehensive dataset of more than 300 automatically generated demonstrations collected in the Isaac Sim environment. As illustrated in Figure~\ref{fig:sim_data_frame} This dataset included synchronized multi-modal observations (RGB, Depth, and proprioception) and action streams, enabling participants to train both visuomotor imitation learning and VLA models effectively. The dataset was designed to cover a wide range of initial conditions to encourage robust policy learning. Note that the data generation scripts were also provided, allowing teams to augment the dataset with additional samples if desired. The dataset, along with the baseline implementations and deployment instructions, was made available on the official competition website to ensure that all participants had a common starting point for their development efforts. \href{https://huggingface.co/datasets/rocochallenge2025/rocochallenge2025/tree/main/gearbox_assembly_demos_updated}{\textcolor{blue}{This link}} provides access to the simulation dataset.

\subsection{Onsite Track: Physical Collaborative Assistance}
The onsite finals took place at the Advanced Remanufacturing and Technology Centre (ARTC) in Singapore. Finalists deployed their models onto the Galaxea R1 Lite, a dual-arm mobile manipulation platform equipped with head-mounted stereo cameras and wrist-mounted RGB-D sensors.

\subsubsection{Hardware and Data Collection}

\begin{figure}[t]
    \centering
    \includegraphics[width=0.9\columnwidth]{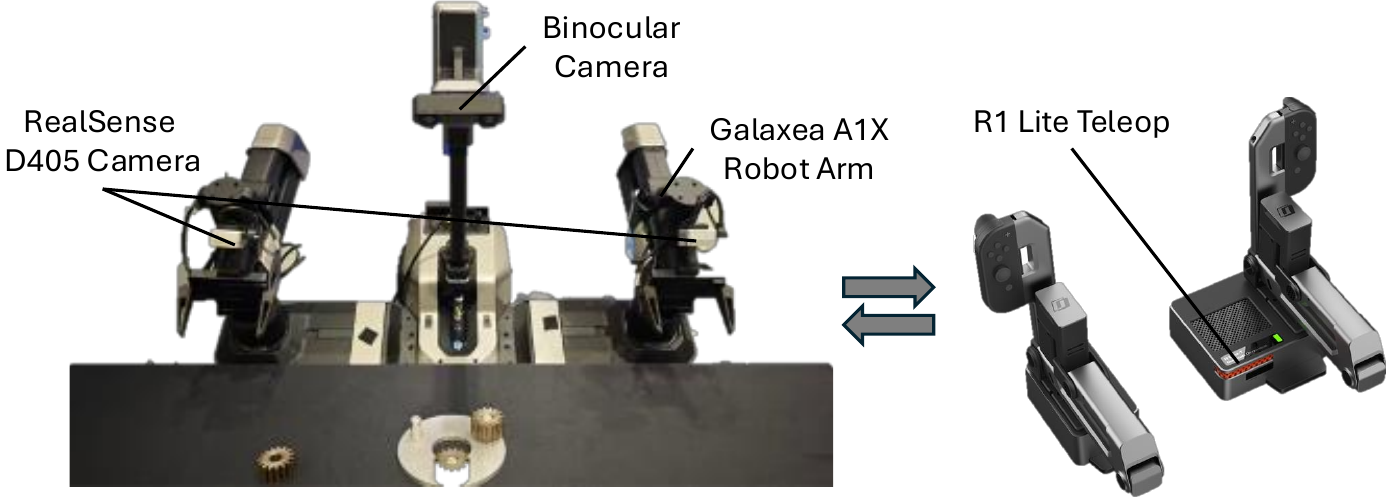}
    \caption{Hardware settings for real-world data collection. The platform consists of a Galaxea A1X robot arm, multiple RealSense D405 cameras, a binocular camera, and an R1 Lite teleoperation device used for demonstration collection.}
    \label{fig:hardware_settings}
\end{figure}

Unlike the simulation phase, participants in the real-world track were provided with a high-quality teleoperated dataset consisting of more than 300 demonstrations collected on the target hardware platform. As shown in Figure~\ref{fig:hardware_settings}, the hardware setup includes a Galaxea A1X robot arm, multiple RealSense D405 cameras, a binocular camera, and an R1 Lite teleoperation device. The dataset contains synchronized multi-modal observations, including RGB images, depth measurements, and proprioceptive robot states, together with corresponding action streams, thereby supporting the training of imitation learning and vision-language-action (VLA) models.

\begin{figure}[t]
     \centering
     \begin{subfigure}[b]{\linewidth}
         \centering
         \includegraphics[width=\linewidth]{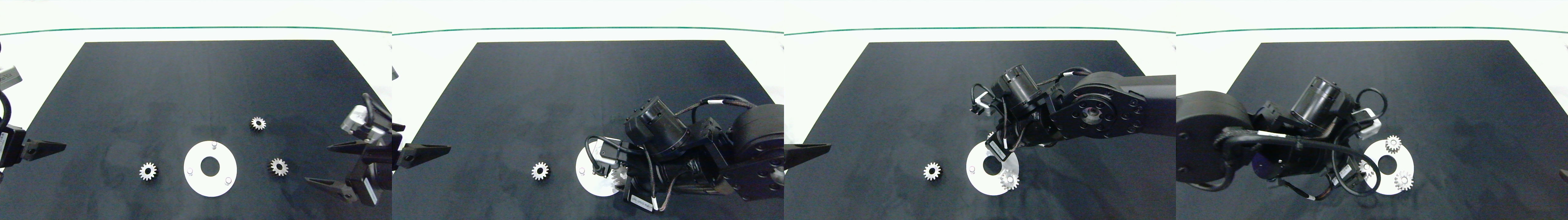}
         \caption{Task 1: Assemble 3 gears}
         \label{fig:real1}
     \end{subfigure}
     \hfill
     \begin{subfigure}[b]{\linewidth}
         \centering
         \includegraphics[width=\linewidth]{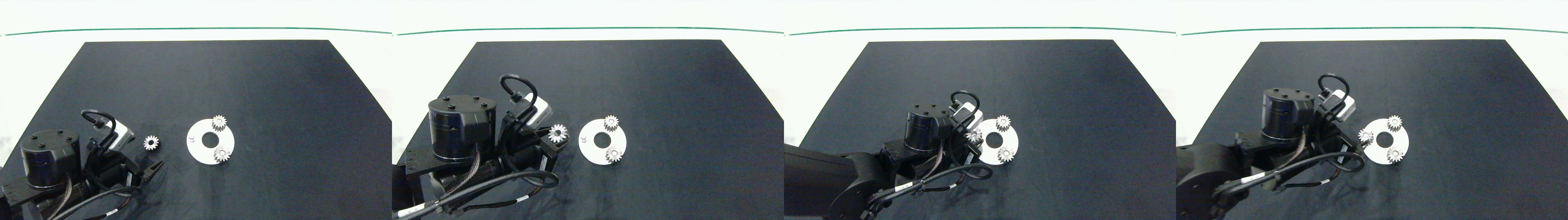}
         \caption{Task 2: Assemble from partial state}
         \label{fig:real2}
     \end{subfigure}
     \hfill
     \begin{subfigure}[b]{\linewidth}
         \centering
         \includegraphics[width=\linewidth]{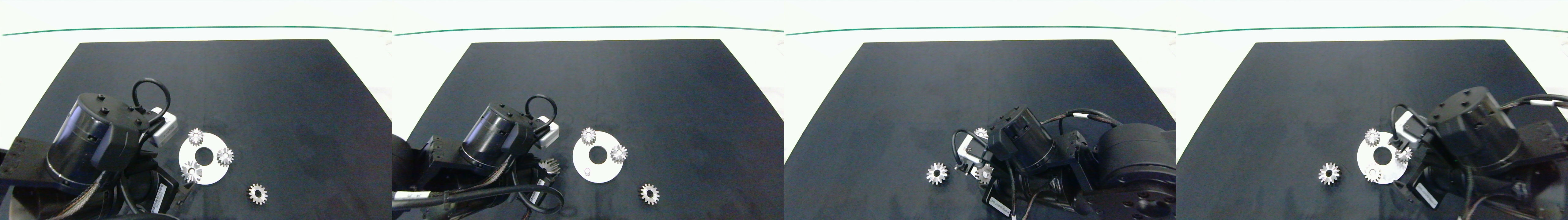}
         \caption{Task 3: Error recovery by replacing the wrong gear}
         \label{fig:real3}
     \end{subfigure}
        \caption{Real-world demonstration of the three gear assembly tasks in the RoCo Challenge. (a) Assembly from scratch; (b) Completion from a partial state; (c) Error recovery by replacing the incorrect gear.}
        \label{fig:real_tasks_all}
\end{figure}

As illustrated in Figure~\ref{fig:real_tasks_all}, the real-world setting is partially different from the simulation track. Task 1 (Figure~\ref{fig:real1}) is simplified to assemble only 3 planet gears: pick them up and insert onto the planet carrier pins one by one. In task 2 (Figure~\ref{fig:real2}) the robot starts from the partial state where two planetary gears are already inserted, and only needs to pick and insert the remaining one. In task 3 (Figure~\ref{fig:real3}) a larger gear is mounted on one pin by mistake, and the robot needs to remove the wrong gear and place the correct one instead.

\begin{figure}
    \centering
    \includegraphics[width=0.8\columnwidth]{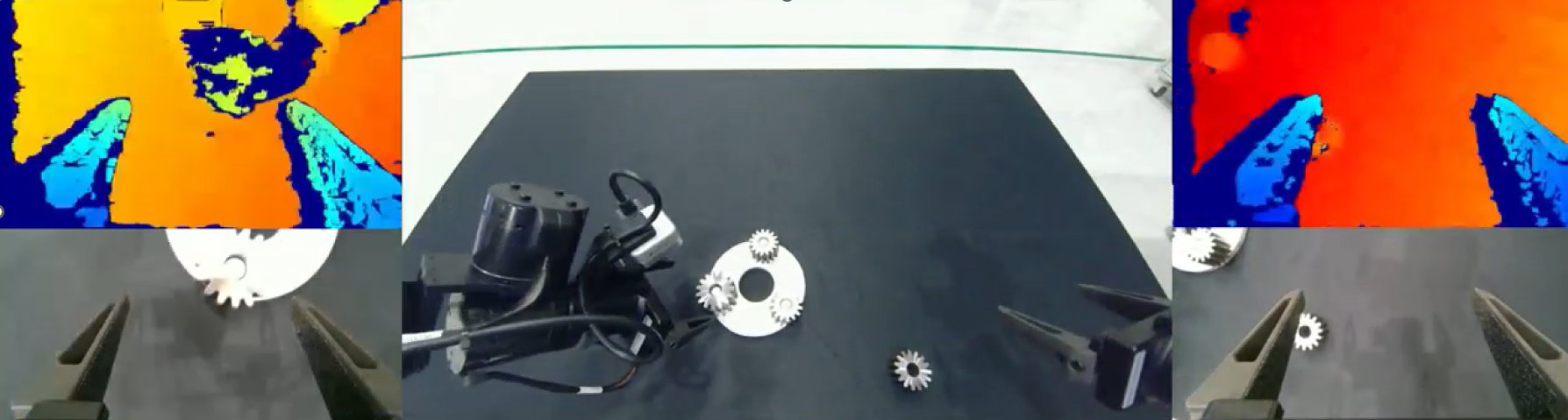}
    \caption{Visualization of the data frame for the real-world track collected using the teleoperation device. The dataset includes synchronized multi-modal observations (RGB, Depth, and proprioception) and action streams, enabling comprehensive training in real-world conditions.}
    \label{fig:real_data_frame}
\end{figure}

\subsubsection{Real-World Constraints}
The physical track introduced complexities absent in simulation, including:
\begin{itemize}
    \item \textbf{Contact Dynamics:} The planetary gears require precise meshing; a misalignment of a few millimeters prevents the sun or ring gear from seating correctly.
    \item \textbf{Sensor Noise:} Teams had to manage the variance in real-world RGB streams and reconstruct depth from stereo head cameras using provided calibration files (if needed).
    \item \textbf{Quick Deployment:} Finalists had only several hours for model deployment and testing on the physical hardware, emphasizing the need for robust sim-to-real transfer and efficient fine-tuning strategies.
\end{itemize}

\subsubsection{Baselines and Data Provision}
Similar to the simulation track, baseline implementations of $\pi_{0.5}$~\citep{pi05} and ACT~\citep{act} were provided for the real-world track, trained on the teleoperated dataset.
For the data collection, we use a teleoperation device (shown in Figure~\ref{fig:hardware_settings}) to collect high-quality demonstrations of the assembly tasks. Noteably, the dataset includes a significant portion of error detection and recovery demonstrations, which are crucial for training models to handle the complexities of real-world assembly. As shown in Figure~\ref{fig:real_data_frame}, the real-world robotic assemby dataset is fully synchronized across multiple modalities, including RGB images from the head-mounted cameras, depth measurements from the wrist-mounted sensors, and proprioceptive data from the robot's joints, which can be accessed through \href{https://huggingface.co/datasets/rocochallenge2025/rocochallenge2025/tree/main/real_assembly_r1lite}{\textcolor{blue}{this link}}.
\subsection{Results Overview}
The competition saw participation from over 60 teams and 170+ participants spanning 10 countries. The results, summarized in Table \ref{tab:results_detailed}, reveal a significant "Sim-to-Real Cliff." While many teams achieved high scores in the controlled simulation environment, only a few transitioned successfully to the physical hardware.

\begin{table}[t]
\centering
\caption{RoCo Challenge Final Leaderboard (Top 6) - Detailed Scores}
\label{tab:results_detailed}
\resizebox{\textwidth}{!}{%
\begin{tabular}{lcccccccccc}
\toprule
 & \multicolumn{4}{c}{\textbf{Simulation Track}} & \multicolumn{4}{c}{\textbf{Real-World Track}} & \\
\cmidrule(lr){2-5} \cmidrule(lr){6-9}
\textbf{Team} & \textbf{$S_1$} & \textbf{$S_2$} & \textbf{$S_3$} & \textbf{Sim Score} & \textbf{$S_1$} & \textbf{$S_2$} & \textbf{$S_3$} & \textbf{Real Score} & \textbf{Final Score} \\
\midrule
\textbf{IIGroup}        & \cellcolor{cellpurple!40}0.190 & \cellcolor{cellblue!25}0.100 & \cellcolor{cellblue!20}0.075 & \cellcolor{cellgreen!30}0.126 & \cellcolor{cellpurple!25}0.111 & \cellcolor{cellblue!60}0.250 & \cellcolor{cellblue!100}\textbf{0.625} & \cellcolor{cellgreen!85}\textbf{0.344} & \textbf{0.279} \\
\textbf{RoboCola}       & \cellcolor{cellpurple!15}0.070 & \cellcolor{cellblue!12}0.050 & \cellcolor{cellblue!0}0.000 & \cellcolor{cellgreen!10}0.038 & \cellcolor{cellpurple!35}\textbf{0.139} & \cellcolor{cellblue!90}\textbf{0.417} & \cellcolor{cellblue!0}0.000 & \cellcolor{cellgreen!35}0.139 & 0.109 \\
\textbf{Show Me Robot}  & \cellcolor{cellpurple!18}0.080 & \cellcolor{cellblue!6}0.025 & \cellcolor{cellblue!0}0.000 & \cellcolor{cellgreen!9}0.037 & \cellcolor{cellpurple!15}0.056 & \cellcolor{cellblue!40}0.167 & \cellcolor{cellblue!25}0.167 & \cellcolor{cellgreen!30}0.122 & 0.097 \\
\textbf{HD-Robo}        & \cellcolor{cellpurple!100}\textbf{0.440} & \cellcolor{cellblue!85}0.400 & \cellcolor{cellblue!0}0.000 & \cellcolor{cellgreen!65}\textbf{0.256} & \cellcolor{cellpurple!0}0.000 & \cellcolor{cellblue!0}0.000 & \cellcolor{cellblue!0}0.000 & \cellcolor{cellgreen!0}0.000 & 0.077 \\
\textbf{K-Lee-gends}    & \cellcolor{cellpurple!25}0.110 & \cellcolor{cellblue!100}\textbf{0.450} & \cellcolor{cellblue!18}\textbf{0.125} & \cellcolor{cellgreen!45}0.184 & \cellcolor{cellpurple!0}0.000 & \cellcolor{cellblue!0}0.000 & \cellcolor{cellblue!0}0.000 & \cellcolor{cellgreen!0}0.000 & 0.055 \\
\textbf{Real2RealGap}   & \cellcolor{cellpurple!22}0.100 & \cellcolor{cellblue!0}0.000 & \cellcolor{cellblue!0}0.000 & \cellcolor{cellgreen!10}0.040 & \cellcolor{cellpurple!0}0.000 & \cellcolor{cellblue!0}0.000 & \cellcolor{cellblue!0}0.000 & \cellcolor{cellgreen!0}0.000 & 0.012 \\
\bottomrule
\end{tabular}%
}
\end{table}

There are several key observations from the results:
\begin{itemize}
    \item \textbf{The Simulation Leaders:} Teams like HD-Robo and K-Lee-gends dominated the simulation round by leveraging modular perception-action pipelines. However, these solutions proved brittle onsite, failing to generalize to the physical robot's latency and mechanical tolerances.
    \item \textbf{The VLA Advantage:} The top three overall winners (IIGroup, RoboCola, and Show Me Robot) all utilized end-to-end VLA architectures. These models demonstrated superior robustness to the domain shift between simulation and reality, particularly in Task 3 (Error Recovery), where semantic reasoning was paramount.
\end{itemize}

\section{Excellent VLA Solutions}
The top-performing teams in the RoCo Challenge primarily leveraged end-to-end VLA architectures. Unlike modular pipelines that rely on brittle geometric chains, VLA models demonstrated a superior ability to internalize the complex relationship between visual semantics and motor control, making them more resilient to the "Sim-to-Real" gap. This section details the technical innovations of the two leading teams: IIGroup (1st Place Overall) and RoboCola (2nd Place Overall).
\subsection{ARC-VLA and Failure-Aware Data Programming}
\begin{figure*}[t]
    \centering
    \includegraphics[width=0.8\textwidth]{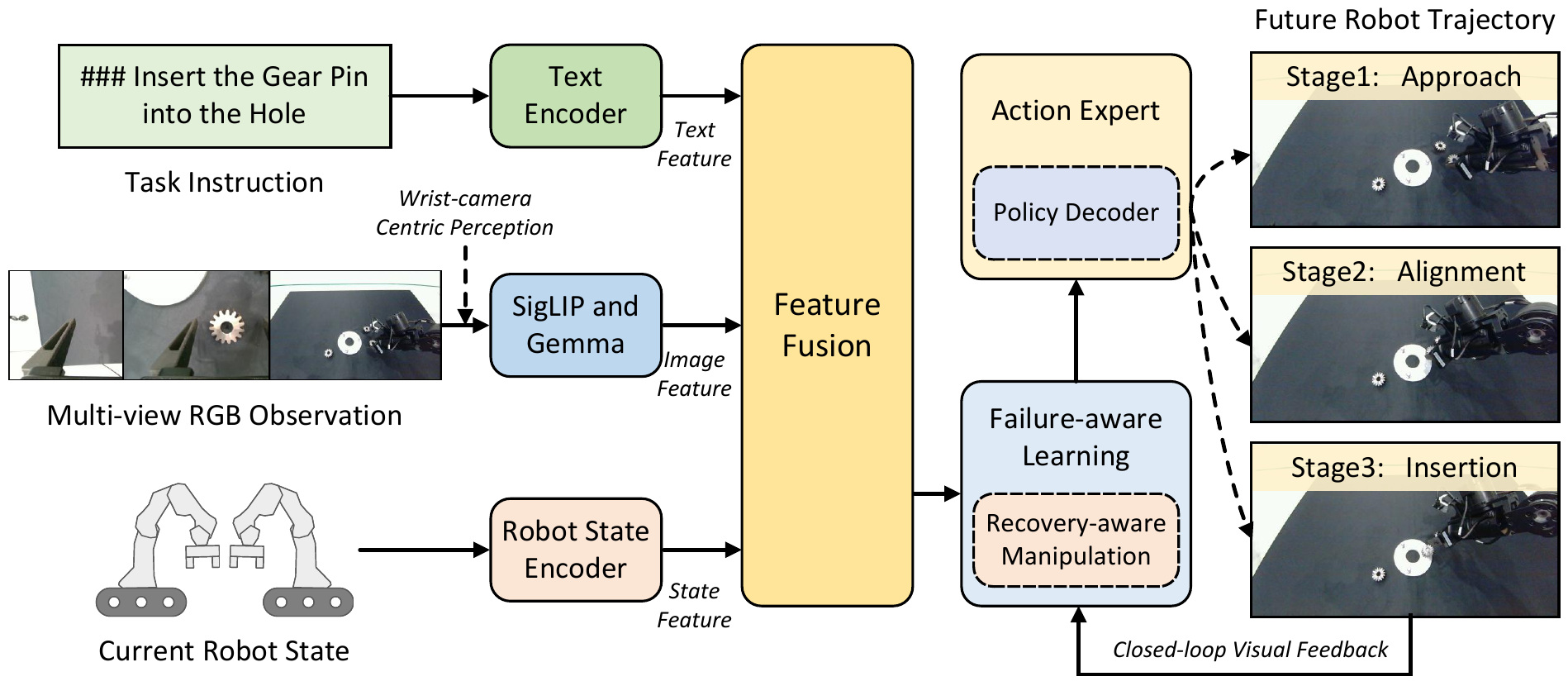}
    \caption{Overall pipeline of ARC-VLA. The model integrates task instruction, multi-camera observations, and robot state features through a multi-modal backbone and a failure-aware policy learning framework to generate closed-loop manipulation actions.}
    \label{fig:arc_vla_pipeline}
\end{figure*}

The overall winner, IIGroup, introduced ARC-VLA, a 3.6-billion-parameter model built upon the $\pi_{0.5}$ backbone~\citep{pi05}. As illustrated in Figure~\ref{fig:arc_vla_pipeline}, their success was largely attributed to a stability-first design philosophy that emphasized consistent observation protocols and explicit failure recovery. Instead of relying solely on nominal successful demonstrations, ARC-VLA was developed to maintain robust execution under fine-grained misalignment and repeated trial failures, making it particularly effective for precision-sensitive manipulation tasks. To provide a visual overview of the system design.

\subsubsection{Architecture and "Risk-First" Perception}
ARC-VLA utilizes a hybrid vision backbone combining SigLIP and Gemma with a dedicated action expert~\citep{siglip, gemma}. To circumvent the complexities of global depth calibration, the team adopted a wrist-camera-centric strategy. Wrist cameras provide high-fidelity "near-field" data of the gear and pins, allowing the model to learn contact geometry implicitly without relying on noisy global depth maps.

\subsubsection{Overcoming the "Lazy Robot" with L1 Loss}
A key technical insight from IIGroup was addressing the precision bottleneck. They observed that standard L2 (MSE) loss leads to vanishing gradients as the robot approaches a target, causing it to stop correcting when it is "close enough" but not yet aligned for insertion. To combat this, they implemented L1 loss optimization, which maintains a constant gradient, forcing the model to minimize even sub-millimeter errors:
$$ \mathcal{L}_{L1} = \sum_{t=1}^{T} |\mathbf{a}_{pred}^{(t)} - \mathbf{a}_{gt}^{(t)}|.$$
They maintained constant gradient pressure, enabling millimeter-level insertion accuracy. Furthermore, they heavily utilized recovery-from-failure curriculum data, constructing specific datasets targeting misalignment recovery and infinite-loop breaking, which proved critical for Task 3 success.
\subsubsection{Recovery Correction Data Programming}
Instead of training only on "perfect" trajectories, IIGroup explicitly taught the robot how to manage failures through a two-stage training pipeline (600 simulation episodes, 400 real-world episodes). They introduced two specific error types:
\begin{itemize}
    \item \textbf{Type 1 (Small Offsets):} Correcting minor misalignments during placement.

    \item \textbf{Type 2 (Fail and Reset):} Teaching the robot to completely retract, recalibrate, and retry after a failed grasp or a "stuck" pin scenario.
\end{itemize}
\subsubsection{Deployment Alignment}
To stabilize the policy onsite, they reduced the action chunking size $K$ from 50 to 20. This allowed the robot to react more dynamically to visual feedback (closed-loop control) rather than blindly following a long, potentially erroneous trajectory predicted from a single frame.
\subsection{RoboCola}
\begin{figure*}[!t]
    \centering
    \includegraphics[width=\textwidth]{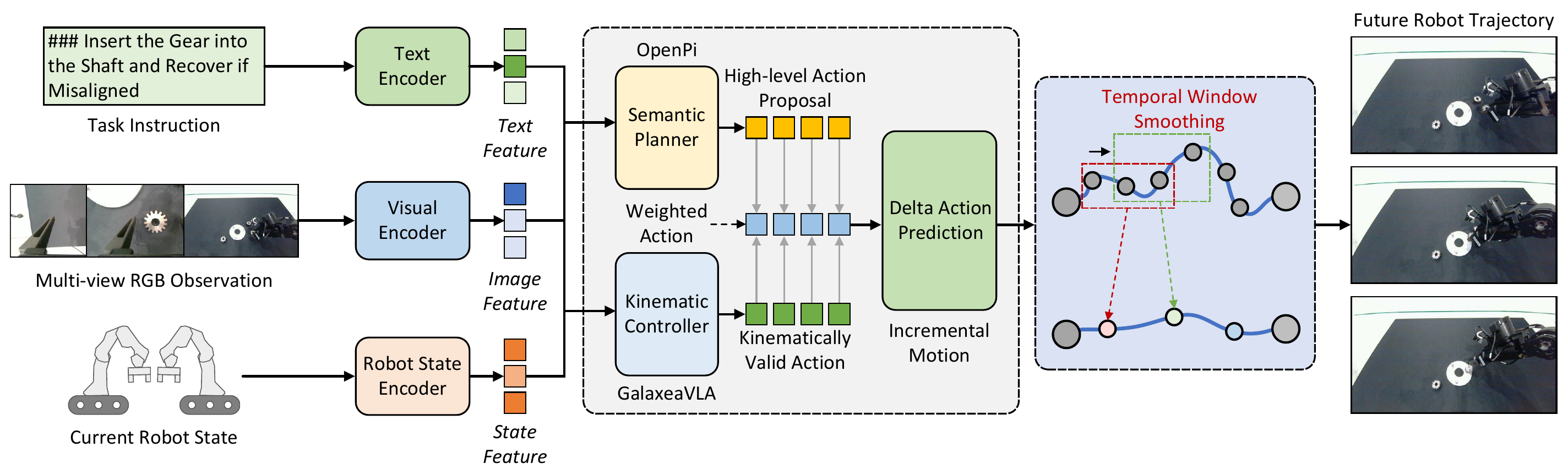}
    \caption{Overall pipeline of the Team RoboCola. The framework adopts a dual-model collaboration strategy to couple high-level task reasoning with low-level motion generation, enabling effective and precise robotic manipulation.}
    \label{fig:robocola_pipeline}
\end{figure*}

Team RoboCola secured the runner-up position by employing a unique Dual-Model Collaboration Framework that balances high-level logical reasoning with low-level kinematic precision. As illustrated in Figure~\ref{fig:robocola_pipeline}, the overall system is organized as a collaborative pipeline in which different model components are responsible for complementary aspects of decision-making and control. This design enables RoboCola to combine semantic reasoning at the task level with precise action execution at the motion level, leading to strong performance in complex manipulation scenarios.

\subsubsection{The Dual-Model Framework}
Securing the runner-up position, RoboCola effectively demonstrated a Specialist + Generalist dual-model framework. Their approach fused outputs from a semantic planner (OpenPi) and a kinematic controller (GalaxeaVLA)~\citep{pi05, galaxeavla}. The final action is a weighted combination prioritizing kinematic precision.
RoboCola argued that a single model struggles to handle both fine-grained motor skills and complex assembly logic (e.g., Task 3 recovery). Their solution fused two models:
\begin{itemize}
    \item \textbf{Generalist (OpenPi):} Acted as the "high-level brain" to maintain task logic and handle sequential error correction.
    \item \textbf{Specialist (GalaxeaVLA):} A model pre-trained on the specific kinematics of the Galaxea R1 Lite to ensure smooth, physically valid movements.The final action was a weighted fusion: $$ \mathbf{a}_{final} = \lambda \cdot \mathbf{a}_{spec} + (1-\lambda) \cdot \mathbf{a}_{gen}. $$
\end{itemize}

\subsubsection{Delta Action Space and Trajectory Smoothing}
To preserve sensitivity for micro-adjustments, RoboCola utilized a Delta Action Space rather than absolute coordinates. This prevented the model from "losing sensation" during the final millimeters of insertion. To reduce jitter, they applied a temporal window algorithm to smooth transitions between consecutive action chunks, preventing high-frequency oscillations that would otherwise make gear meshing physically impossible.

\subsubsection{Rejection Sampling Fine-Tuning (RFT)}
To bridge the data gap for failure recovery, the team used RFT. They allowed the model to practice in the Isaac Sim simulator, collected all successful attempts (including those where the model accidentally recovered from a mistake), and added these trajectories back into the training set. This "self-correction" loop significantly boosted performance in Task 1 and Task 2. High-quality simulated rollouts were filtered by a success function $R(\tau)$ and fed back into the training dataset:
$$ \mathcal{D}_{new} = \mathcal{D}_{old} \cup \{ \tau \sim \pi_\theta | R(\tau) = 1 \}. $$

\subsubsection{Inference Optimization}
During the onsite finals, RoboCola identified a "sweet spot" of 7 execution steps per inference cycle. This balanced the trade-off between visual lag (too many steps) and computational stuttering (too few steps), maintaining fluid motion on the provided RTX 6000 Ada hardware.

\section{Excellent Perception-Action Solutions}

While VLA models dominated the real-world rankings, modular Perception-Action paradigms demonstrated remarkable efficiency and precision in the simulation phase. These approaches decompose the end-to-end mapping into explicit stages of state estimation, planning, and control. This section analyzes the strategies of HD-Robo and K-Lee-gends, who achieved the highest scores in the Simulation Track.

\subsection{Task Decomposition and Rule-Based Twisting}

Team HD-Robo secured the highest score in simulation by treating the gearbox assembly as a series of decoupled geometric problems. Their pipeline was designed for modular reliability, though it faced significant challenges during the physical transfer.
The team utilized a combination of Foundation Pose estimation and depth-based thresholding. For the initial three gears, the system identified parts using semantic segmentation and calculated 6D poses relative to the planet carrier pins.
Recognizing that precise gear meshing is difficult to achieve via pure position control, HD-Robo implemented an ad-hoc twisting logic. Once the gear reached the vicinity of the pin, the robot executed a "Twist-Left, Twist-Right" motion primitive. This rule-based approach allowed the gears to "settle" into the meshed state by searching for local geometric minima, effectively bypassing the need for complex contact modeling.
Despite its simulation success, HD-Robo’s reliance on Franka-style impedance control rules, which were not natively supported by the Galaxea R1 Lite’s control interface, led to a deployment failure onsite. This underscores the difficulty of transferring modular policies that are tightly coupled to specific simulation-based physics parameters.

\subsection{Geometric Priors and Spiral Search}
Team K-Lee-gends adopted a purely geometric approach, focusing on robust perception and deterministic motion primitives to handle the uncertainties of assembly.
To handle the transition from simulation textures to real-world lighting, the team replaced their initial Mask R-CNN~\citep{maskrcnn} with the Segment Anything Model 3 (SAM 3)~\citep{sam3}. This allowed for high-fidelity part segmentation across varying environmental conditions without requiring extensive real-world re-training.
To address the "hole-searching" problem in gear-to-pin assembly, the team implemented a Spiral Search algorithm. When the end-effector encountered a vertical resistance force ($F_z$) exceeding a predefined threshold, the robot would transition from a linear move to a planar spiral trajectory:
\begin{equation}
x(t) = r(t) \cos(\omega t), \quad y(t) = r(t) \sin(\omega t)
\end{equation}
where the radius $r(t)$ increases linearly over time. This allowed the robot to find the pin location even under slight calibration offsets.
The team utilized MegaPose~\citep{megapose} for object pose estimation and a traditional Inverse Kinematics (IK) solver for motion generation. While this provided high precision in the noise-free simulation, the cumulative errors in the physical perception-action loop, compounded by system latency, prevented successful execution on the physical hardware.

\section{Key Insights}

The RoCo Challenge provided a unique lens into the current state of EAI for industrial automation. The stark contrast between simulation rankings and real-world performance offers several critical insights for the research community.

\subsection{The "Sim-to-Real Cliff" and Paradigm Robustness}

A defining takeaway from the competition was the performance gap between modular and end-to-end systems. Modular pipelines dominated the Simulation Track, where they benefited from perfect state estimation and zero-latency control. However, these systems proved brittle onsite, failing primarily due to:

\begin{itemize}
    \item \textbf{Calibration Sensitivity:} Small errors in camera intrinsics or joint offsets caused cascading failures in geometric chains.
    \item \textbf{Latency Jitter:} Real-world communication delays between perception and control modules often led to unstable oscillations during contact-rich tasks.
\end{itemize}

In contrast, VLA-based approaches demonstrated higher tolerance for sensor noise. By learning a direct mapping from pixels to actions, these models internalize calibration offsets as part of the visual distribution, making them far more resilient to the physical realities of the assembly bench.

\subsection{Data Engineering vs. Model Scale}

The success of ARC-VLA highlights that for high-precision industrial tasks, data quality and diversity are more decisive than raw parameter count. While many teams focused on increasing model size, the winning strategy relied on:

\begin{itemize}
    \item \textbf{Failure-Recovery Curriculum:} Training on data that explicitly includes errors, such as "stuck" pins or dropped parts, transformed the policies from "hope-for-success" models into resilient agents.
    \item \textbf{Loss Function Alignment:} Replacing standard $L2$ loss with $L1$ loss was a critical technical pivot. By maintaining constant gradients, $L1$ loss prevents the "lazy robot" phenomenon, ensuring that the agent continues to apply micro-corrections until the assembly is fully seated.
\end{itemize}

\subsection{The Bottleneck of Visual Occlusion}

Task 1 (Assembly from Scratch) revealed a fundamental hardware-software limitation: visual occlusion. As the robot's hand approaches the carrier pin, the gear itself often blocks the camera's view of the target.

\begin{itemize}
    \item \textbf{The Wrist-Camera Solution:} Teams that utilized wrist-mounted cameras (IIGroup) had a significant advantage in handling near-field geometry.
    \item \textbf{Multi-Arm Collaboration:} Future solutions may require "active perception," where one arm holds the part while the other arm positions its wrist camera to provide a secondary "side-view" for alignment, effectively solving the occlusion problem through bimanual coordination.
\end{itemize}

\subsection{Semantic Reasoning in Manipulation}

Task 3 (Error Detection and Recovery) proved to be the most difficult scenario. It required the robot to not only perform a motor skill but also to understand the semantic state of the workspace (e.g., "This gear is the wrong size"). Only the VLA-based models, which leverage the reasoning capabilities of Large Language Models (LLMs), were able to successfully identify and correct these high-level errors. This suggests that the future of industrial robotics lies in the fusion of low-level "reflexive" motor control with high-level semantic reasoning.
\section{Competition Impact}

The RoCo Challenge has established a new benchmark for evaluating EAI within the context of HCM. Beyond the technical scores, the competition served as a catalyst for community building and a rigorous testbed for the "Sim-to-Real" transition in industrial automation.

\subsection{Global Participation and Community Reach}

The challenge attracted significant international interest, drawing more than 60 teams and 170+ participants from over 10 countries. This diversity brought together leading academic institutions, such as Tsinghua University, Beihang University, NUS, and GIST, alongside innovative industrial players like HiDream.ai. This cross-sector collaboration facilitated a unique exchange of ideas between foundational AI research and practical industrial engineering.

\subsection{Onsite Engagement and Knowledge Sharing}

The onsite finals, held at the A*STAR ARTC in Singapore, provided a high-visibility platform for state-of-the-art robotics. The event culminated in a dedicated session at the International Workshop on Addressing Challenges and Opportunities in Human-Centric Manufacturing at AAAI 2026.

\begin{itemize}
    \item \textbf{Live Audience:} The onsite challenge attracted approximately 100 attendees, providing a rare opportunity for researchers to observe the real-world failure modes and successes of various algorithmic paradigms in person.
    \item \textbf{Technical Plenary:} The top-performing teams, IIGroup and RoboCola, delivered invited talks detailing their system architectures. These presentations offered invaluable "lessons from the trenches," specifically regarding the management of real-world noise and the implementation of recovery strategies.
\end{itemize}

\subsection{Standardizing Industrial Benchmarks}

By providing a standardized hardware platform (Galaxea R1 Lite) and a unified evaluation kit, the RoCo Challenge addressed a major bottleneck in robotics research: reproducibility. The release of the "RoCo-Dataset", which includes over 300 high-quality teleoperated demonstrations and multi-modal sensor logs, provides the community with a persistent resource for training and benchmarking future VLA and modular models.

The challenge has effectively shifted the conversation from \textit{can a robot perform a task in simulation?} to \textit{how robustly can a robot handle the uncertainty of a physical assembly line?} In doing so, it has set a new standard for what constitutes a "deployable" EAI solution in the manufacturing sector.

\section{Related Work}

The RoCo Challenge builds upon and extends several foundational research areas within the robotics community, bridging the gap between theoretical robot learning and practical industrial application.

\subsection{Robotic Manipulation for Industrial Automation}

Traditional industrial automation has long relied on Task and Motion Planning (TAMP) and rigid programming for high-speed assembly~\citep{tampsurvey, manureview}. While these systems excel in structured environments, they often fail when faced with the variability of modern "Human-Centric Manufacturing." Recent benchmarks, such as the NIST Assembly Task Board, have sought to standardize the evaluation of robotic dexterity~\citep{nist}. The RoCo Challenge advances this by introducing collaborative assembly, where robots must not only possess dexterity but also the semantic reasoning to resume tasks from partial states or recover from human-induced errors.

\subsection{Robot Learning in Manipulation}

The transition from modular pipelines to end-to-end learning has been a dominant trend in recent years~\citep{eaisurvey, rt2}.

\begin{itemize}
    \item \textbf{Visuomotor Policies:} Approaches like ACT and Diffusion Policy (DP) have demonstrated remarkable success in learning complex, multi-modal tasks from demonstrations~\citep{act, dp}.
    \item \textbf{VLA Models:} Models such as RT-2, $\pi_{0.5}$, and OpenVLA have introduced the concept of "Embodied Foundation Models," leveraging LLM weights to provide semantic priors for manipulation~\citep{rt2, pi05, openvla}.
\end{itemize}

The RoCo Challenge specifically benchmarks these models against the high-precision requirements of gearbox assembly, highlighting that while VLA models offer superior reasoning, they still require specialized fine-tuning, such as the $L1$ loss and failure-recovery data utilized by the winning teams, to meet industrial standards.

\subsection{Long-Horizon Manipulation and Multi-Task Learning}

Assembling a planetary gearbox is a quintessential long-horizon task, requiring dozens of successful sub-actions where errors can compound over time~\citep{act}. Research into Hierarchical Reinforcement Learning (HRL) and Sub-goal Discovery has attempted to solve these sequences by breaking them into manageable primitives~\citep{hrl, sgd}. The RoCo Challenge demonstrates a shift toward unified, task-conditioned policies that handle diverse scenarios (scratch, resume, recovery) within a single architecture, proving that multi-task learning can enhance the robustness of individual skills through shared representations.

\section{Future Plan}

The 1st RoCo Challenge has provided a foundational benchmark, but it also revealed several avenues for further research and development to fully realize the potential of EAI in smart manufacturing. Our future efforts will focus on three primary dimensions:

\subsection{Expanding the Simulation Ecosystem}

While the current Isaac Sim-based environment effectively modeled the planetary gearbox assembly, we plan to expand the simulator to support a broader range of complex industrial tasks. This includes:

\begin{itemize}
    \item \textbf{Industrial Product Inspection:} Integrating high-fidelity visual and surface defect simulation to train agents for quality control.
    \item \textbf{PCB Sorting and Assembly:} Introducing small-scale, high-precision electronic component handling that requires even greater dexterity and specialized end-effectors.
    \item \textbf{Dynamic Human-Robot Collaboration:} Moving beyond "virtual human" state initialization to real-time, physics-based human models to test reactive safety and proactive assistance in shared workspaces.
\end{itemize}

\subsection{Advancing Algorithmic Robustness}

To address the "Sim-to-Real Cliff" identified in this challenge, future research will explore:

\begin{itemize}
    \item \textbf{Autonomous Recovery Discovery:} Utilizing Reinforcement Learning (RL) within the simulator to autonomously discover optimal recovery policies, reducing the reliance on manually curated failure datasets.
    \item \textbf{Temporal Memory Integration:} Incorporating architectural features like Transformers or State Space Models (SSMs) with longer temporal windows to prevent "infinite loop" failures and allow the robot to learn from its own recent history of attempts.
    \item \textbf{Active Bimanual Perception:} Developing policies that coordinate both arms, one for manipulation and the other for positioning a "third-eye" camera, to overcome visual occlusion during contact-rich tasks.
\end{itemize}

\subsection{Real-World Scaling}

We aim to transition from standardized kits to more varied, real-world industrial components. This involves testing models on parts with different materials (e.g., reflective metals, flexible plastics) and under the non-idealized lighting conditions typical of active factory floors.

\section{Conclusion}

The 1st RoCo Challenge successfully benchmarked the state-of-the-art in robotic collaborative manipulation for industrial assembly. By attracting over 60 teams and 170+ participants, the competition highlighted the global research community's commitment to bridging the gap between EAI research and manufacturing reality.

Our analysis of the top-performing solutions underscores a significant shift in the field. While modular perception-action pipelines showed promise in simulation, VLA models emerged as the clear winners in the physical world. Their ability to handle sensor noise and perform semantic reasoning for error recovery (Task 3) suggests that end-to-end learning, when grounded with physical priors like $L1$ loss and failure-recovery data programming, is the most promising path toward deployable industrial agents.

The challenge has set a new standard for industrial benchmarks, providing the community with the RoCo-Dataset and a rigorous evaluation framework. As we move forward, the insights gained here, specifically regarding the importance of data engineering over model scale and the need for active perception, will serve as a roadmap for the development of "human-centric" robots capable of working seamlessly alongside human operators in the factories of the future.

\section*{Acknowledgment}

We acknowledge the support of the Singapore National Robotics Programme, A*STAR ARTC, and Galaxea for providing the hardware and venue support.

\bibliographystyle{plainnat}
\bibliography{references}

\end{document}